\pdfoutput=1

\documentclass[11pt]{article}

\usepackage{acl}

\usepackage{times}
\usepackage{latexsym}

\usepackage[T1]{fontenc}

\usepackage[utf8]{inputenc}

\usepackage{microtype}
\usepackage{latexsym}
\usepackage{microtype}
\usepackage{subfigure}
\usepackage{amssymb}
\usepackage{stfloats}
\usepackage{algorithm}
\usepackage{algorithmic}
\usepackage{float}
\usepackage{graphicx}
\usepackage{amsmath}
\usepackage{booktabs}
\usepackage{hyperref}

\title{Hierarchical information matters: Text classification \\ 
via tree based graph neural network}

\author{Chong Zhang$^{*}$, He Zhu$^{*}$, Xingyu Peng, Junran Wu$^{\dag}$, Ke Xu$^{\dag}$ \\
        State Key Lab of Software Development Environment, Beihang University, Beijing, 100191, China\\
        \texttt{\{chongzh, Roy\_Zh, xypeng, wu\_junran, kexu\}@buaa.edu.cn}}

\begin{document}
\maketitle

\begin{abstract}
Text classification is a primary task in natural language processing (NLP). Recently, graph neural networks (GNNs) have developed rapidly and been applied to text classification tasks. As a special kind of graph data, the tree has a simpler data structure and can provide rich hierarchical information for text classification. Inspired by the structural entropy, we construct the coding tree of the graph by minimizing the structural entropy and propose HINT, which aims to make full use of the hierarchical information contained in the text for the task of text classification. Specifically, we first establish a dependency parsing graph for each text. Then we designed a structural entropy minimization algorithm to decode the key information in the graph and convert each graph to its corresponding coding tree. Based on the hierarchical structure of the coding tree, the representation of the entire graph is obtained by updating the representation of non-leaf nodes in the coding tree layer by layer. Finally, we present the effectiveness of hierarchical information in text classification. Experimental results show that HINT outperforms the state-of-the-art methods on popular benchmarks while having a simple structure and few parameters.
\end{abstract}

\renewcommand{\thefootnote}{}
\footnotetext{$^{*}$Equal Contribution.}
\footnotetext{$^{\dag}$Correspondence to: Junran Wu, Ke Xu.}

\section{Introduction}

Text classification is an essential problem in NLP. There are numerous applications of text classification, such as news filtering, opinion analysis, spam detection, and document organization~\cite{TCAP}. Recently, GNNs have developed rapidly. GNNs learn the representation of each node by aggregating the information of neighboring nodes and can retain structural information in the graph embedding. Therefore, many graph-based methods are applied to text classification and achieve good performance.~\cite{TextGCN} proposed TextGCN, which is the first method to employ a Graph Convolutional Network (GCN) in the text classification task. They built a heterogeneous graph containing word nodes and document nodes for the corpus and transformed the text classification task into a node classification task. TextGCN outperformed other traditional methods and attracted much attention, which has led to increasingly more applications of graph-based methods in text classification. In~\citet{huang2019,ING}, the text classification task was converted into the graph classification task. They built text-level co-occurrence graphs for each data.~\citet{huang2019} employed a Message Passing Mechanism (MPM) and outperformed TextGCN. A Gated Graph Neural Network (GGNN) was employed in~\cite{ING} and achieved state-of-the-art performance.

\begin{figure*}[!ht]
\centering
\includegraphics[width=\textwidth]{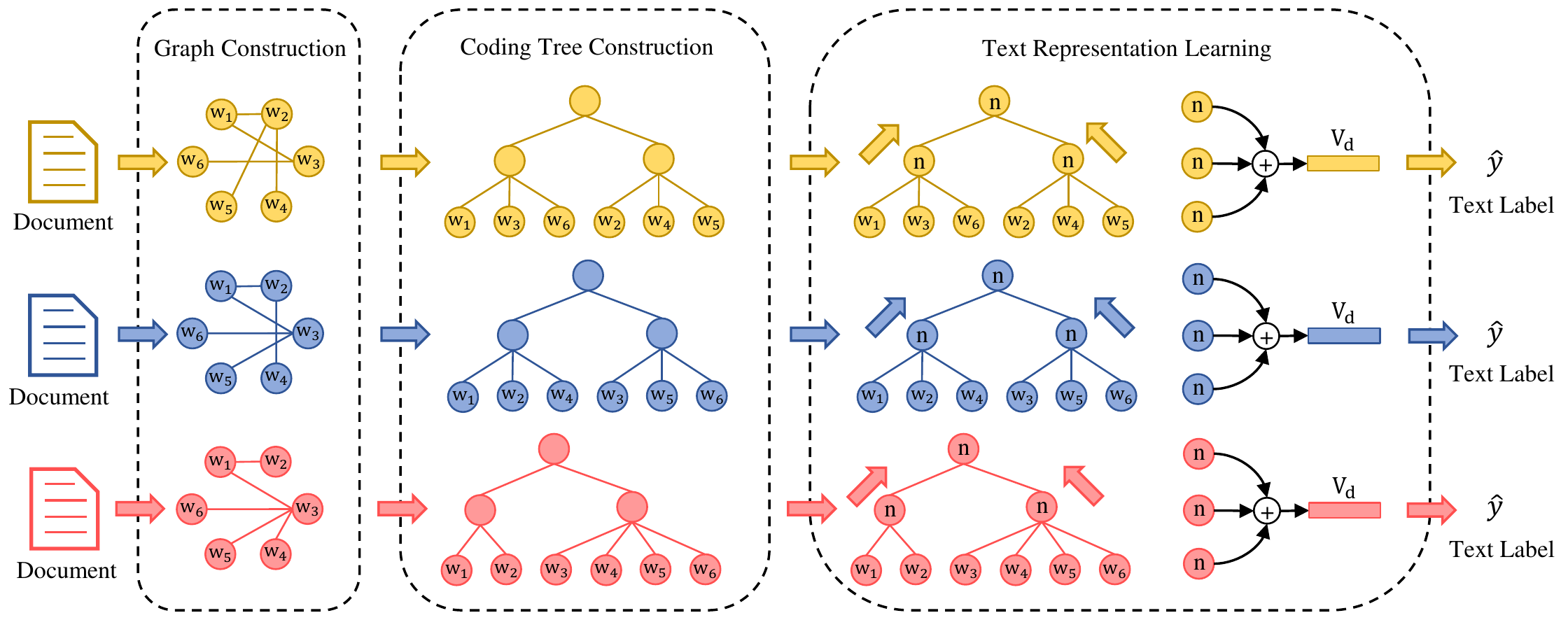}
\caption{\label{fig:model}The architecture of HINT. $w_{i}$ nodes represent the word nodes, others are non-leaf nodes in the coding tree.}
\end{figure*}
Different from normal graph data, tree-structured data is a simpler data structure with rich hierarchical information. In the text classification task, despite plenty of efforts that have been devoted to the adoption of GNNs, none of them has realized the rich hierarchical information in text, which has already been employed in other NLP tasks. TrDec was proposed for the NMT task to generate a target-side tree topology and uses the tree to help the translation process~\cite{TreeNMT}. In knowledge-based question answering tasks,~\citet{TreeKBQA} treated the query as a tree and used the tree-based LSTM to model the context of the entities or relationships in the query.~\citet{Treehelp2} clustered the labels in the extreme multi-label text classification task and built a label tree.~\citet{treehelp4} proposed a rumor detection model based on tree transformer to better exploit user interactions in the dialogues. Hierarchical information in text is likely to be helpful for text classification tasks, so it should be better utilized.

Inspired by structural entropy~\cite{SI,wu2022structural,wu2022simple}, we propose a novel model based on tree structure for text classification, named HINT. Structural entropy can measure the complexity of the hierarchical information of the graph, and decode its key structure. As shown in Figure~\ref{fig:model}, we first build individual graph for each document through dependency parsing. Then the graphs are transformed into their corresponding coding trees by a structural entropy minimization algorithm. The coding tree not only retains the crucial features of data but also excludes many other features that worsen the model. The model classifies the entire document by learning the representation of the coding tree. So far, we not only make better use of the hierarchical information in the text data but also represent text with a simpler data structure (i.e., tree). We conduct several experiments to verify the advantages of our method over the baselines. To sum up, our contributions are as follows:
\begin{itemize}
\item For the first time, we explore the effectiveness of hierarchical information of documents in text classification.
\item We propose a novel method, HINT, which aims to parse and represent the hierarchical information of documents.
\item The results demonstrate that our method not only outperforms several text classification baselines but is also much simpler in structure than other graph-based models.
\end{itemize}

\section{Related Work}
In recent years, the graph-based text classification method transforms the text classification task into a graph classification task or a node classification task and has achieved good performance in the text classification task. Different from traditional deep learning models~\cite{CNN,LSTM}, the graph-based text classification methods usually capture the rich relationships between nodes in the graph by constructing document graphs or corpus graphs, then apply GNN to learn the embedding of the document, and finally input the embedding to the classification layer. For graph construction, one is to construct a static graph for text or corpus. Some methods are to construct a single heterogeneous graph for the entire corpus~\cite{TextGCN,GCN2}. Other studies construct a separate graph for each document to handle the inductive learning setting~\cite{huang2019,ING}. Except for static graphs, the construction of dynamic graphs does not rely on prior knowledge and can be jointly learned with the model~\cite{DYgraph}. For the learning of graph representation, various GNN models are used in text classification, such as GCN~\cite{TextGCN,GCN2}, GGNN~\cite{ING}, MPM~\cite{huang2019}, and GAT-based model~\cite{GAT}. 

Tree structure data is a special kind of graph data with a simple structure and rich hierarchical information. With the development of deep learning, many models use trees to help solve NLP tasks. Some methods process the data to get the tree structure data and use it to help the model.~\cite{TreeNMT} utilized the strong correlation between grammatical information and tree structure, and a tree-based decoder TrDec is proposed for the NMT task. TrDec generates a target-side tree topology and uses the tree to guide the translation process.~\cite{Treehelp2} proposed a model based on label tree and attention mechanism for extreme multi-label text classification. The label tree is constructed by clustering the labels and the model is trained from top to bottom. In addition to using the generated tree, the deep learning model of the tree structure is also widely used. Tree-structured multi-linear principal component analysis (TMPCA) used the PCA of each layer to transfer the representation of two adjacent words to the next layer until the entire text is reduced to a single-word vector~\cite{Treehelp3}.~\cite{TreeKBQA} treated the query as a tree and proposed a tree-based LSTM in the Knowledge-based question answering task to model the context of entities or relationships in the query.

The tree structure can provide hierarchical information for the model, and the tree based deep learning model can make better use of the grammatical information through the hierarchy of the model. However, hierarchical information is not well studied in graph-based text classification methods. In this work, we aim to fill the gap in GNN-based text classification.

\section{Method}
In this section, we will introduce HINT in detail. First, we will explain the method of constructing a graph for each text. Then, we introduce the coding tree construction algorithm that can decode the hierarchical information in text data. Finally, we will show how the model learns the hierarchical information and text representation based on tree-structured data and how to predict the label for a given text based on the learned representations. 

\subsection{Graph Construction}
In previous graph-based text classification models, there are two ways to construct a graph for text. In~\cite{TextGCN}, the corpus is constructed into a heterogeneous graph containing word nodes and document nodes. The weight of the edge between nodes is the point-wise mutual information (PMI) of the words or the TF-IDF value. This method can explicitly model the global word co-occurrence and can easily adapt to graph convolution. The other method is to construct a graph for each text.~\citet{huang2019,ING} construct a graph for a textual document by representing unique words as vertices and co-occurrences between words as edges. This method reduces memory consumption and is friendly to new text.

However, the graphs constructed by the above methods do not contain rich syntax and semantics information. In addition, the method based on co-occurrence treats the words at different positions equally, which causes a lack of position information in the representations of graph nodes. To retain more features in the constructed graphs, we use dependency parsing to construct a graph in HINT. We perform dependency parsing on each sentence in the document to obtain the dependencies between words. In addition, the dependency parsing result of each sentence contains a root word. We connect the root words of adjacent sentences to form a complete dependency parsing graph. 

Take a document with $l$ words $D=\{w_1,\ldots,w_i, \ldots,w_l\}$, where $w_i$ is the $i_{th}$ word of document. The set of dependencies between words is $DP=\{dr_{ij}|i\neq j; i,j\leq l\}$, where $dr_{ij}$ denotes the dependency relation of words $i$ and $j$. The edges between pairs of words with dependencies is $E_w=\{e_{ij}|dr_{ij} \in DP\}$. The root word set of each sentence is $DR=\{r_i|i\leq n\}$, where $n$ represents the number of sentences in the document. $E_r=\{e_{ij}|r_i,r_j\in DR\land j=i+1\}$ represents the set of edges between the root words of adjacent sentences. The dependency parsing graph $G=(V, E)$ for a text is defined as:
\begin{align}
V &= \{w_i| i\in [1,l]\}, \\
E &= \{E_w \cup E_r\},
\end{align}
where $V$ and $E$ are the node set and edge set of the graph respectively. Dependency parsing can analyze the semantic associations between words. Using words as nodes and dependency relationships as edges, the graph has rich semantic and structural information.

\subsection{Coding Tree Construction}
After the text is transformed into graph structure data through dependency analysis, each graph has rich structural information. In this subsection, we introduce our method to decode the hierarchical information in text data from constructed graphs. 

In~\cite{SI}, the structural entropy of the graph is defined as the average amount of information of the codewords obtained by a random walk in a specific coding pattern. According to the different coding patterns, structural entropy can measure the dynamic information of the graph. Given a graph $G=(V, E)$, the structural entropy of $G$ on partitioning tree $T_p$ is defined as:
\begin{equation}
H^{T_p}(G)=-\sum_{\alpha \in T_p} \frac{g_{\alpha}}{2m}\log{\frac{V_{\alpha}}{V_{\alpha^-}}},
\end{equation}
where $m=|E|$, $\alpha$ is the non-root node of $T_g$ and represents a subset of $V$, $\alpha^-$ is the parent of $\alpha$, $g_{\alpha}$ represents the number of edges with only one end point in $\alpha$ and the other end outside $\alpha$, $V_{\alpha}$ and $V_{\alpha^-}$ is the sum of the degree of nodes in $\alpha$ and $\alpha^-$. The structural entropy of $G$ is defined by $H(G)=min_{T_p}{H^{T_p}(G)}$. $T_p$ is also called the coding tree. Coding tree is simpler form of data, while retaining key features of the original graph. For a certain coding mode, the height of the coding tree should be fixed. Therefore, the h-dimensional structural entropy of the graph $G$ determined by the coding tree $T$ with a certain height $h$ can be computed as:
\begin{equation}
H^{h}(G)=\min_{\{T|h(T) \leq h\}}H^{T}(G).
\end{equation}

We designed a graph coding algorithm based on minimizing structural entropy to transform the graph to its coding tree. The coding tree construction algorithm shown in Algorithm~\ref{alg:algorithm} is based on the principle of minimizing structural entropy to construct a h-dimensional coding tree with a certain height $h$. So the coding tree $T$ with given height $h$ is computed by $T = SEMA(G, h)$, where $T=(V_T)$, $V_T = (V^0_T,\dots,V^h_T)$ and $V^0_T = V$. $SEMA$ refers to the structural entropy minimization algorithm. In $SEMA$, the graph is first transformed into a full-height binary coding tree, and then the tree is folded into a coding tree with certain height $h$. 

\begin{figure}[!ht]
\centering
\includegraphics[width=0.415\textwidth]{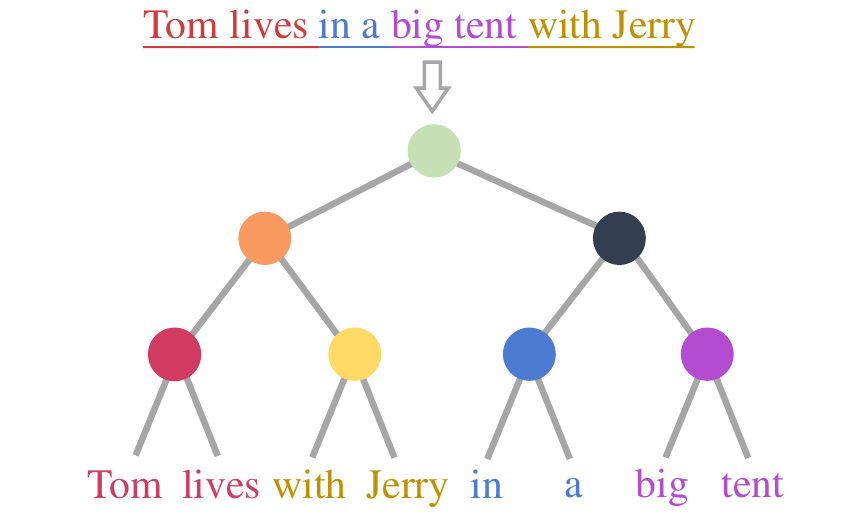}
\caption{\label{fig:exampletree}A text and its corresponding coding tree with a height of 3. The green node represents the root node of the coding tree, and other points refer to other inner nodes with various hierarchical information.}
\end{figure}

\renewcommand{\thealgorithm}{\arabic{algorithm}}
\begin{algorithm}[!htb]
\caption{Structural Entropy Minimization Algorithm}
\label{alg:algorithm}
\textbf{Input}: Adjacency matrix $A_{m*m}$ of the graph $G=\{V,E\}$ \\
\textbf{Parameter}: The height $h$ of the coding tree \\
\textbf{Output}: The coding tree $T=\{V_T,E_T\}$ of the graph $G$ 
\begin{algorithmic}[1] 
\STATE Let $heap=[], V_T=V, unmerge=m, M=\varnothing $.\\
\FOR{$n_i, n_j$ in $V$ and $E_{ij}\in E$}
\STATE $T^{'}= T.Combine(n_i, n_j)$
\STATE $\Delta e = entropy(T) - entropy(T^{'}) $
\STATE $heap.push(\Delta e, n_i, n_j)$
\ENDFOR

\WHILE{$unmerge > 1$}
\STATE $n_1, n_2 = heap.pop(min\Delta e)$
\IF {$n_1 \notin M$ and $n_2 \notin M$}
\STATE $n_{new} = merge(n_1, n_2)$
\STATE update $V_T, M, unmerge$
\STATE update $heap$ like line 4 to 6
\ENDIF
\ENDWHILE
\STATE $root=n_{new}$
\STATE update $heap$ like line 4 to 6
\WHILE{the height of the tree $> h$}
\STATE $n_p, n_c = heap.pop(min\Delta e)$
\STATE $n_p= compress(n_p, n_c)$
\STATE $V_T.del(n_c)$
\ENDWHILE
\STATE \textbf{return} $T$
\end{algorithmic}
\end{algorithm}

For a text and its corresponding coding tree, all words in the text are the leaf nodes, and the hierarchical information in text are decoded into the hierarchical structure of the coding tree. Figure~\ref{fig:exampletree} shows an example of a text and its corresponding coding tree. We can see that the coding tree divides the words of the text into four parts. The information contained in each non-leaf node can be interpreted as the semantic information of all its child nodes, so non-leaf nodes at different levels contain semantic information with different granularities. In addition, the two words "Tom" and "Jerry" that are farther apart in the original text are closer in the coding tree while retaining the correct semantic information.

\subsection{Text representation learning}
Based on the decoded hierarchical structure of coding trees, we aim to learning the text representation with this hierarchical information. Specifically, following the message passing mechanism in GNNs, we intend to iteratively update the node representation of coding tree from leaves to root node. Finally, model can obtain a representation of the text by using the structure of the coding tree and the features of the leaf nodes. The number of layers of the model is the same as the height of the coding tree. The $i_{th}$ layer on coding tree $T=(V_T,E_T)$ can be expressed as:
\begin{equation}
x_v^i=MLP^i(\sum\nolimits_{n\in C(v)} x_n^{i-1}),
\end{equation}
where $v\in V_T$, $x_v^i$ is the feature vector of node $v$ with height $i$, $x^0$ is the word embeddings, and $C(v)$ is the child nodes of $v$. The coding tree learning model starts from the leaf node layer and learns the representation of each node layer by layer until reaching the root node. Finally, all feature vectors of the nodes are used to compute a representation of the entire coding tree $x_T$:
\begin{equation}
\begin{split}
x_T=Concat(Pool(\{x_v^i|v\in V_T^i\})\\|i=0,1,2,\dots,h)), 
\end{split}
\label{equ:treerep}
\end{equation}
where $x_v^i$ is the feature vector of node $v$ with height $i$ in $T$, and $h$ is the height of $T$. $Pool$ in Equation~\ref{equ:treerep} can be replaced with a summation or averaging function. In the beginning, the non-leaf nodes have no representation, and the representation of the non-leaf nodes is updated as the information is propagated from the lower layers to the upper layers along the edges of the coding tree. In the process of propagation, the textual information from the leaf nodes interacts with the hierarchical information abstracted by the coding tree, so that the representation of the final document can contain more useful information. Existing methods have fully exploited the local information of text, and our model takes the global information of text into consideration by combining the hierarchical structure of text.

By converting the graph into a coding tree, the data structure becomes simpler, and the hierarchical information and main features of the graph are retained in the coding tree. In HINT, the node feature vector is aggregated in one direction owing to the hierarchical structure of the coding tree, which suggests that our learning model is simple and its convergence is strong. We take the coding trees $\mathbb{T}=(T_1,T_2,\dots,T_n)$ and their feature matrices as inputs. Each word is represented by the GloVe~\cite{Glove} vector and one-hot position encoding vector, and we concatenate these two vectors as feature matrix $F$. The representation of the entire graph can be obtained from Equation~\ref{equ:treerep}, and then the predicted label of the original text is computed as:
\begin{equation}
y_i = softmax(Wx_{T_i}+b),
\end{equation}
where $T_i\in \mathbb{T}$, $x_{T_i}$ is the representation of coding tree $T_i$; and $W$ and $b$ are the weight and bias, respectively. The goal of training is to minimize the cross-entropy between the ground truth label and predicted label:
\begin{equation}
loss = -\sum_{i} g_i\log(y_i),
\end{equation}
where $g_i$ is the one-hot vector of the ground truth label.

\section{Experiments}

\renewcommand{\thefootnote}{\arabic{footnote}}

In this section, we evaluate the effectiveness of HINT\footnote{The code of HINT can be found at \url{https://github.com/Daisean/HINT}.} and report the experimental results.

\paragraph{Datasets.} We utilize datasets including R8, R52, MR, and Ohsumed. R8 and R52 are subsets of the Reuters 21578 dataset. MR is a movie review dataset used for sentiment classification in which each review is a single sentence. The Ohsumed corpus, which is designed for multilabel classification, is from the MEDLINE database. In this paper, we only use single-label data like other GNN-based text classification models~\cite{TextGCN,huang2019}. We employ StanfordNLP~\cite{stanfordnlp} to build the dependency graphs for all datasets. The statistics of our datasets are summarized in Table~\ref{tab:datasta}.

\begin{table}[!htp]
\centering
\resizebox{0.48\textwidth}{!}{
\begin{tabular}{l|c|c|c|c}
 \hline
 \textbf{Datasets} & \# \textbf{Training} & \# \textbf{Test} & \textbf{Categories} & \textbf{Avg. Length} \\ 
 \hline
 MR & 7108 & 3554 & 2 & 20.39  \\ 
 Ohsumed & 3357 & 4043 & 23 & 193.79 \\ 
 R52 & 6532 & 2568 & 52 & 106.29 \\ 
 R8 & 5485 & 2189 & 8 & 98.87 \\ 
 \hline
\end{tabular}}
\caption{Summary statistics of datasets.}
\label{tab:datasta}
\end{table}

\begin{table*}[!htb]
\centering
\begin{tabular}{l|c|c|c|c}
 \hline
 \textbf{Model} & \textbf{MR} & \textbf{R8} & \textbf{R52} & \textbf{Ohsumed} \\ 
 \hline
 CNN(Non-static) & \textbf{77.75}$\pm$ 0.72 & 95.71$\pm$ 0.52 & 87.59$\pm$ 0.48 & 58.44$\pm$ 1.06  \\ 
 RNN(Bi-LSTM) & 77.68$\pm$ 0.86 & 96.31$\pm$ 0.33 & 90.54$\pm$ 0.91 & 49.27$\pm$ 1.07 \\ 
 fastText & 75.14$\pm$ 0.20 & 96.13$\pm$ 0.21 & 92.81$\pm$ 0.09 & 57.70$\pm$ 0.49 \\ 
 SWEM & 76.65$\pm$ 0.63 & 95.32$\pm$ 0.26 & 92.94$\pm$ 0.24 & 63.12$\pm$ 0.55 \\ 
 TextGCN & 76.74$\pm$ 0.20 & 97.07$\pm$ 0.10 & 93.56$\pm$ 0.18 & 68.36$\pm$ 0.56 \\ 
 \citet{huang2019} & - & 97.80$\pm$ 0.20 & 94.60$\pm$ 0.30 & \textbf{69.40$\pm$ 0.60} \\ 
 S$^2$GC & 76.70$\pm$ 0.00 & 97.40$\pm$ 0.10 & 94.50$\pm$ 0.20 & 68.50$\pm$ 0.10 \\
 \hline
 HINT & \textbf{77.03$\pm$ 0.12} & \textbf{98.12$\pm$ 0.09} & \textbf{95.02$\pm$ 0.18} & \textbf{68.79$\pm$ 0.12} \\
 \hline
\end{tabular}
\caption{Test accuracy($\%$) of models on text classification datasets. The average standard deviation of our model is reported based on ten runs.}
\label{tab:allresult}
\end{table*}

\paragraph{Baselines.}
In this paper, we aim to address the utilization of hierarchical information in text classification with GNNs; thus, besides a bunch of popular baselines, we mainly select the comparison methods based on GNNs. We divide the baseline models into three categories: (i) traditional deep learning methods, including the CNN and LSTM; (ii) word embedding methods, including fastText~\cite{fasttext} and SWEM~\cite{SWEM}; and (iii) graph-based methods for text classification, including the spectral approach-based TextGCN , S$^2$GC~\cite{SSGC}, and nonspectral method-based~\cite{huang2019}.

\paragraph{Settings.}
We randomly divide the training set into the training set and the validation set at a ratio of 9:1. We use the Adam optimizer with an initial learning rate of 10$^{-3}$ and set the dropout rate to 0.5. The height of the coding tree is between 2 and 12. We set the sum or average function as the initial function of $Pool$ in Equation~\ref{equ:treerep}. For word embedding, we use pretrained GloVe with the dimension of 300, and the out-of-vocabulary (OOV) words are randomly initialized from the uniform distribution [-0.01, 0.01]. For each position, we set this position in the vector to 1 and the rest to 0 to get the position encoding vector. We concatenate the GloVe vector and position encoding vector as the initial representation of graph nodes.

\subsection{Experimental Results}
Table~\ref{tab:allresult} presents the performance of our model and baselines. Graph network-based methods generally outperform other types of methods because of the inclusion of structural information. We can observe that the performance of our model is generally better than those of other graph network-based methods. Traditional deep learning methods (CNN and RNN) perform well on MR dataset with relatively short text lengths but are not as good at processing long text. The word embedding-based methods (fastText and SWEM) use word embedding with contextual information and perform better on the R52 and Ohsumed datasets than traditional deep learning methods.

TextGCN is the first method to apply a graph neural network method in text classification. TextGCN learns the representation of nodes through corpus-level co-occurrence graphs.~\citet{huang2019} uses the co-occurrence window and message passing mechanism to learn the representation of nodes. S$^2$GC is an extension of the Markov Diffusion Kernel used to make the information aggregation in graphs more efficient. The graph learning methods enable each node to learn a better representation by using the information of its farther neighbors and accordingly performs well on all datasets. The HINT model encodes the graph and extracts the key structure through the structure entropy minimization algorithm, and learns on the coding tree to use the hierarchical information to update the node representation. The representations of non-leaf nodes in the coding tree are obtained by a facile method of layer-by-layer updating from the leaf nodes of the coding tree to the root node. The results show that HINT performs better on the MR, R8, and R52 datasets than other graph-based methods and achieves competitive performance on the Ohsumed dataset. Notably, HINT does not have a complicated structure and numerous parameters, but it still generally outperforms other baselines. Next, we will further analyze the height of the coding tree, the coding tree construction algorithm, the comparison with the state-of-the-art methods and the efficiency of the model.

\begin{figure}[!htb]
\centering
\subfigure[MR]{\includegraphics[width=0.2\textwidth]{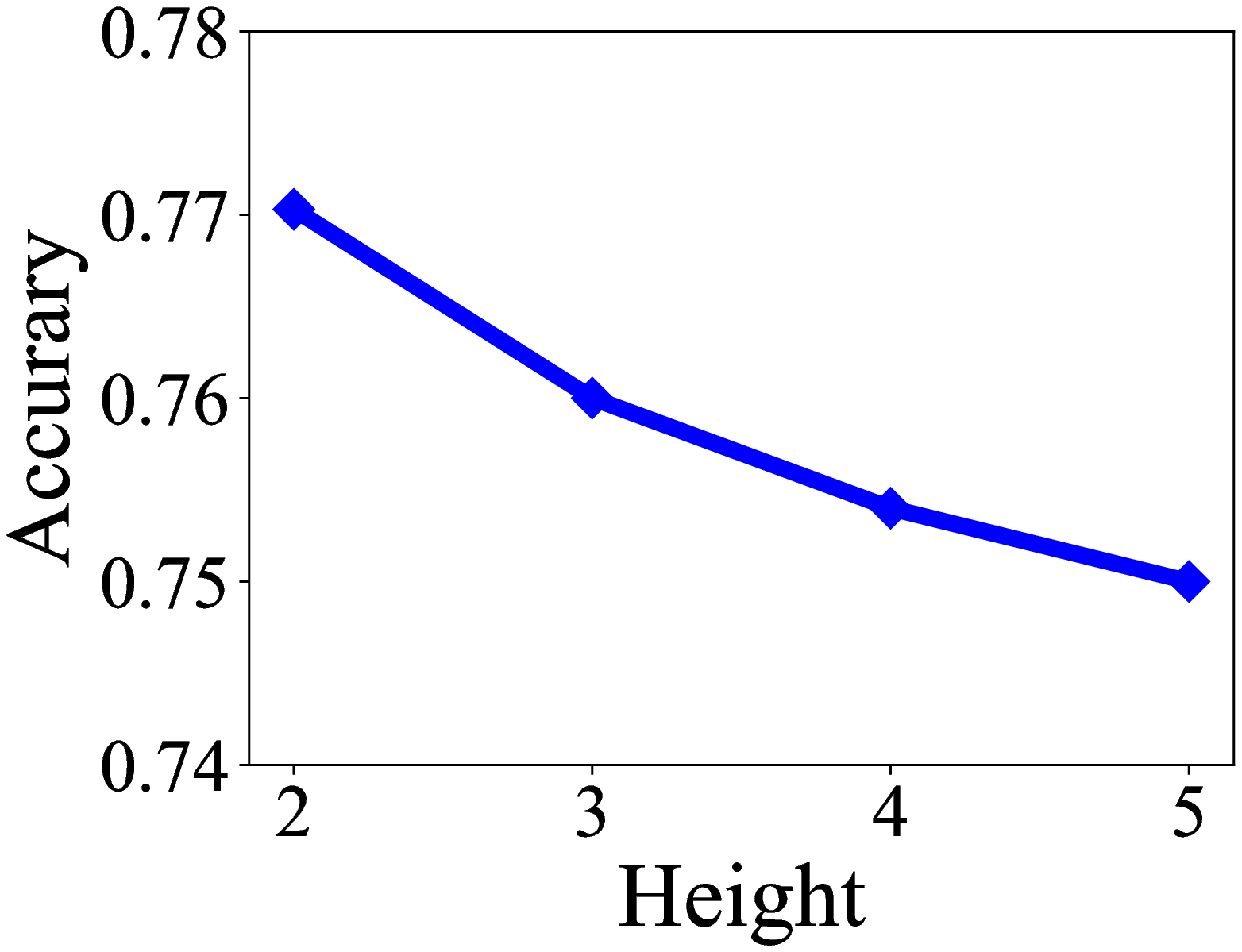}}
\subfigure[R8]{\includegraphics[width=0.2\textwidth]{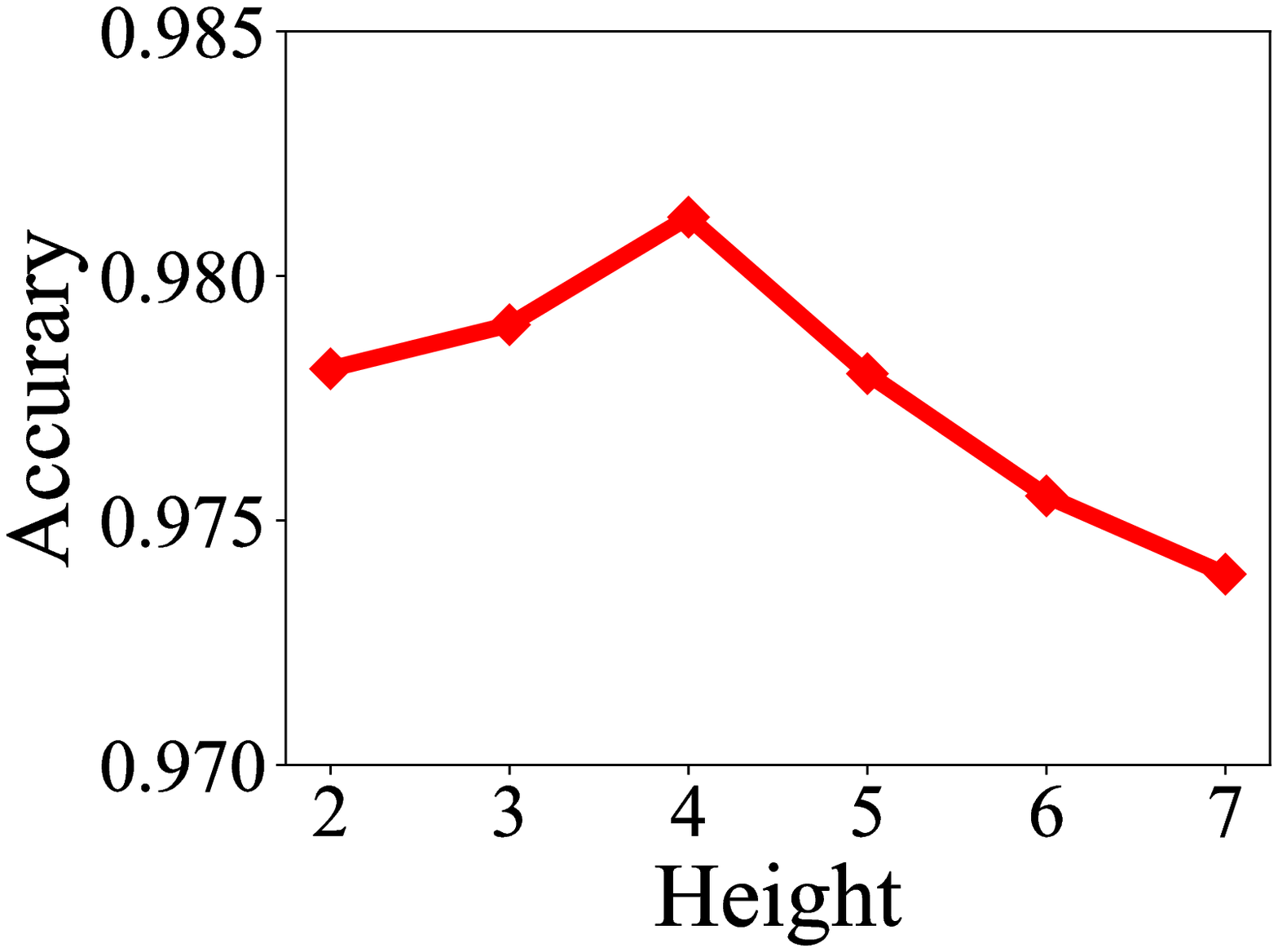}}
\\
\centering
\subfigure[R52]{\includegraphics[width=0.2\textwidth]{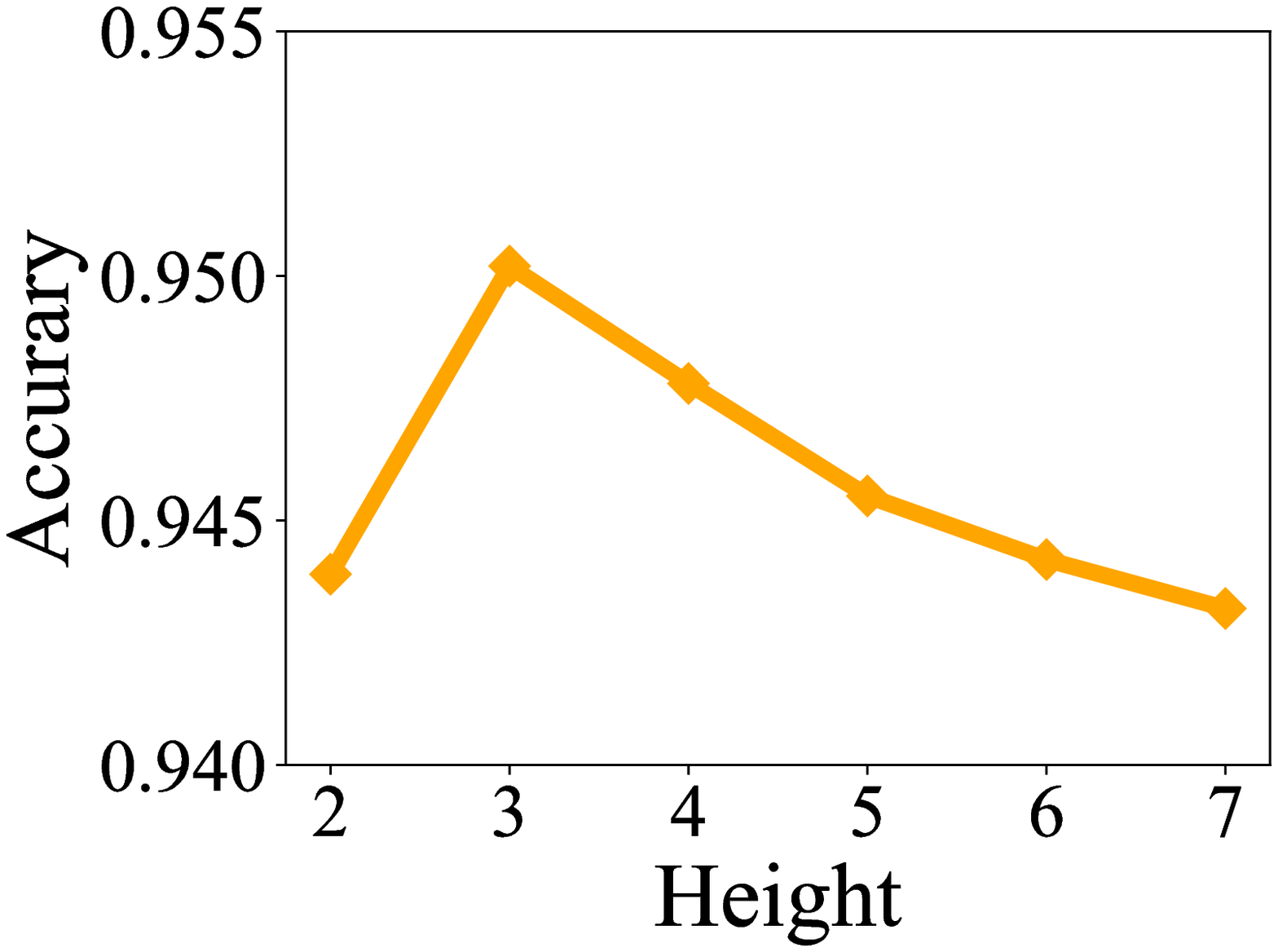}}
\subfigure[Ohsumed]{\includegraphics[width=0.2\textwidth]{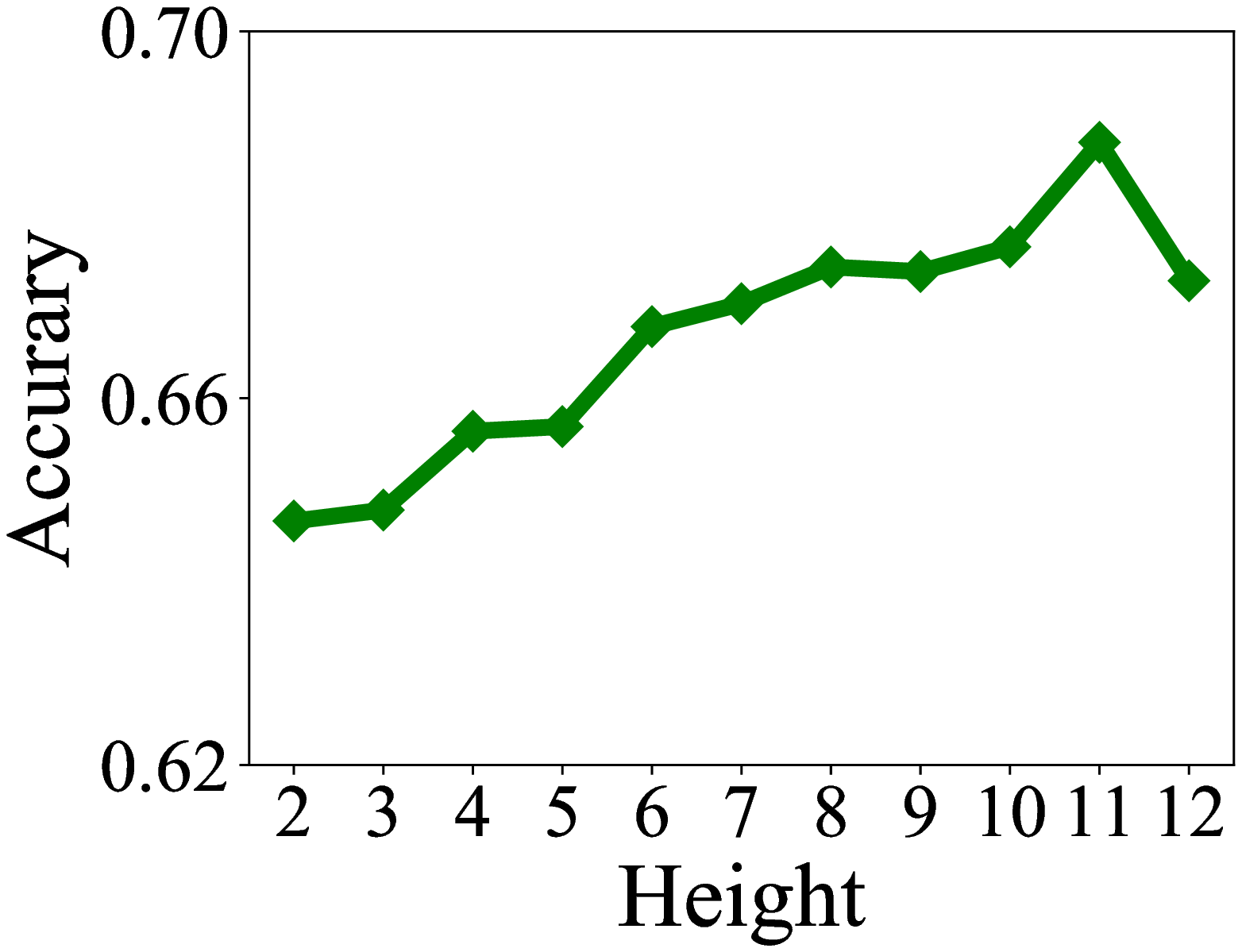}}
\caption{\label{fig:dataheight}The influence of height on the performance of the model on 4 datasets.}
\end{figure}

\subsection{Height of the coding tree}
The height of the coding tree plays an important role in HINT. Different height coding trees reflect divergent hierarchical information, and the utilization of leaf node information is also disparate. Figure~\ref{fig:dataheight} shows the test performance of different height coding trees on the 4 datasets. For datasets with different average lengths, the optimal height of the coding tree is different. Longer texts have more complex structural information, so a higher coding tree is needed to retain this information. In text representation learning, a deeper level is also needed to utilize hierarchical information. For the MR dataset, a height of 2 is the best. For the Ohsumed dataset with an average length of nearly 200, the best performance occurs when the height is 11. Therefore, an appropriate height will improve the quality of the representations and make better use of the original text and hierarchical information.

\subsection{Validity of coding tree construction algorithm}
In this subsection, we compare different coding tree construction methods to illustrate the effectiveness of the coding tree construction algorithm. For a tree of height $h$, we take all nodes of the text dependency analysis graph as leaf nodes of the tree. For each layer of the coding tree, we randomly select two nodes, use a new node as their parent node, and connect all nodes in the ${h-1}_{th}$ layer to a root node (RT). The results are shown in Table~\ref{tab:difftree}. 
\begin{table}[!hb]
\centering
\resizebox{0.48\textwidth}{!}{
\begin{tabular}{l|c|c|c|c}
 \hline
 \textbf{Model} & \textbf{MR} & \textbf{R8} & \textbf{R52} & \textbf{Ohsumed} \\ 
 \hline
 HINT(RT) & 75.85 & 95.84 & 85.86 & 14.84 \\
 HINT & \textbf{77.03} & \textbf{98.12} & \textbf{95.02} & \textbf{68.79} \\
 \hline
\end{tabular}}
\caption{Test accuracy($\%$) of models on text classification datasets with different coding tree construction method. For both construction methods, we use trees of the same height on the same dataset}
\label{tab:difftree}
\end{table}

The results point out that constructing a coding tree by randomly selecting nodes has a negative effect on the model. The difference between the two methods is small on the data set with the shortest average length, but becomes more pronounced as the data set becomes more complex, especially in the Ohsumed dataset. Random selection of nodes destroys the original semantic information, so it is difficult for the model to learn useful features. The coding tree constructed by the minimization structure entropy algorithm can retain the key information in the graph and abstract the hierarchical information of the text. The results demonstrate the effectiveness of minimizing structural entropy for text graphs.

\subsection{Comparison with state-of-the-art method}
With the development of pre-trained language models (PLMs), graph methods based on PLMs have also been applied to text classification tasks. BertGCN\citep{Bertgcn} achieves state-of-the-art performance by combining BERT\citep{bert} with GCN\citep{GCN_ori}. For a fair comparison, we employ the BERT model ($\rm BERT_{base}$) trained in BertGCN, freeze its parameters and input the text into BERT to get the initial node representation. Other experimental settings are consistent with BertGCN. Moreover, because BERT word embeddings contain positional information, we use the text co-occurrence graph like \citet{huang2019}. The result is shown in Table~\ref{tab:bertgcn}.

\begin{table}[!htb]
\centering
\resizebox{0.48\textwidth}{!}{
\begin{tabular}{l|c|c|c|c}
 \hline
 \textbf{Model} & \textbf{MR} & \textbf{R8} & \textbf{R52} & \textbf{Ohsumed} \\ 
 \hline
 BERT & 85.7 & 97.8 & 96.4 & 70.5 \\
 BertGCN & 86.0 & \textbf{98.1} & 96.6 & \textbf{72.8} \\
 \hline
 HINT(BERT) & \textbf{86.4} & \textbf{98.1} & \textbf{96.8} & \textbf{71.2} \\
 \hline
\end{tabular}}
\caption{Test accuracy($\%$) of models on text classification datasets with different position encodings.}
\label{tab:bertgcn}
\end{table}

The results point out that our model outperforms BertGCN on three out of four benchmarks. Because of BertGCN's settings, we truncate some long texts during preprocessing. Our model can decode the hierarchical information of the whole text, but the operation of truncating the text affects the integrity of the text. Ohsumed suffers the most as the dataset with the longest average length, yet we achieve competitive performance nonetheless. BertGCN is based on BERT and GCN, which pay more attention to local information, so incomplete text does not affect performance. On the remaining three datasets with occasional truncation, our model obtains outperformance, indicating that our model can decode hierarchical information and focus on global features, further demonstrating the positive effect of hierarchical information on text classification tasks.

In general, our model achieves superior performance. With the more informative BERT word embeddings, HINT propagates the information in a bottom-up manner and obtains a better text representation. In the light of the accuracies achieved with BERT, our method shows excellent collaboration ability with large-scale pre-trained language models.

\subsection{The Efficiency of HINT}
\begin{figure}[!htb]
\centering
\includegraphics[width=0.48\textwidth]{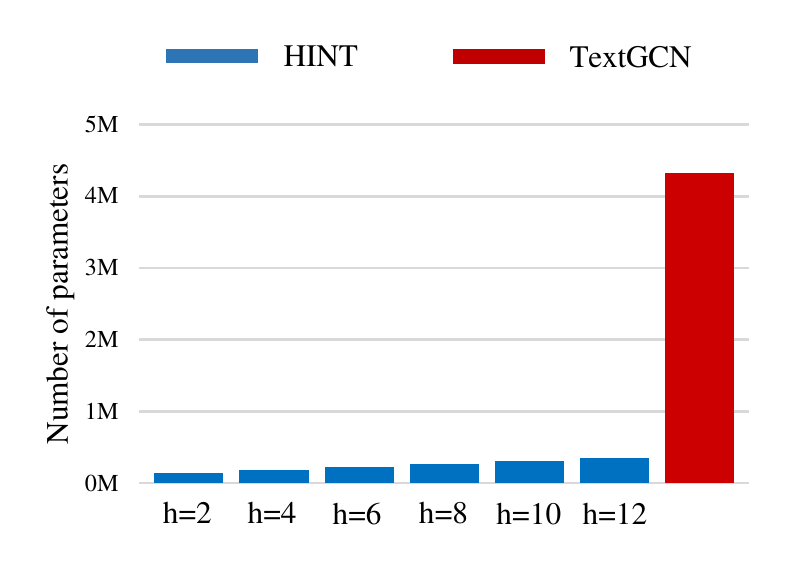}
\caption{\label{fig:para}Comparison of the parameters of HINT and TextGCN on the Ohsumed dataset.}
\end{figure}

In the graph-based baseline models, the computational complexity of TextGCN, S$^2$GC and~\citet{huang2019} is $O(hm)$, where $h$ is the number of diffusion steps and $m$ is the number of edges. The computational complexity of learning model in HINT is $O(n)$, where $n$ is the number of nodes, which is much smaller than that of graph-based baseline models. In addition, we also compare the parameters and floating-point operations per second (FLOPs) of the models. Since S$^2$GC and~\citet{huang2019} do not have a complete model code implementation, we only compare the proposed model with TextGCN. Figure~\ref{fig:para} shows the comparison of the parameters of HINT and TextGCN. We set the height of the coding tree $\in [2,12]$ and use 2 as the step size and then run HINT and TextGCN on the same dataset. We can observe that the parameters of HINT gradually increase as the height increases, but the model with the most parameters is still dozens of times smaller than TextGCN.

\begin{figure}[!htb]
\centering
\includegraphics[width=0.48\textwidth]{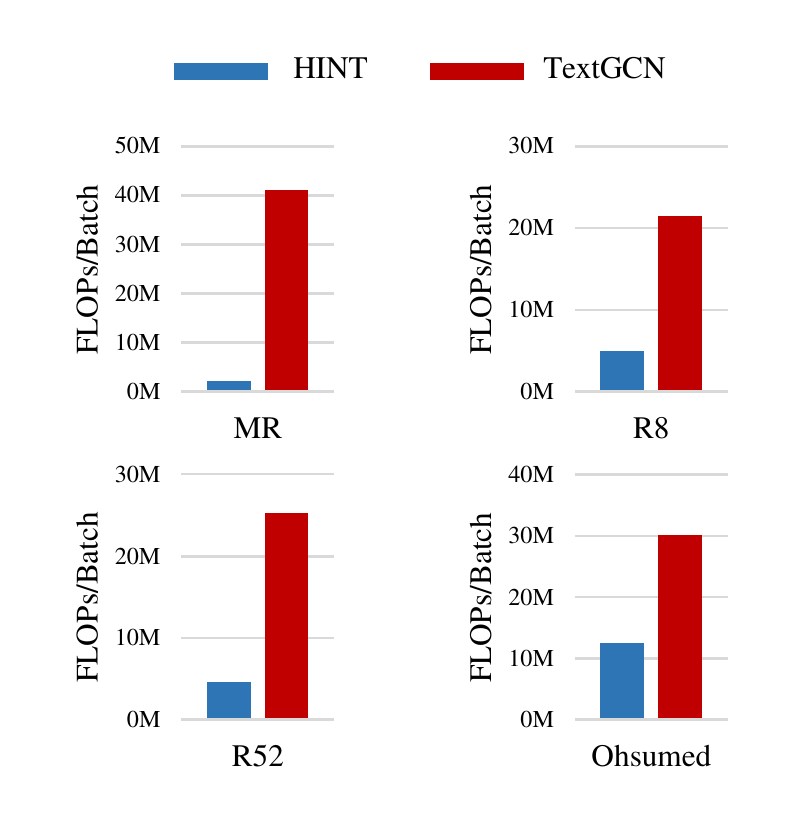}
\caption{\label{fig:flops}Comparison of the FLOPs of HINT and TextGCN.}
\end{figure}

Moreover, we further compare the number of FLOPs of HINT and TextGCN on the same parameter settings. We set the hidden size of HINT to 96 and the batch size to 4. We calculate the FLOPs of the HINT and TextGCN models on four datasets. The results are shown in Figure~\ref{fig:flops}. We can see that the calculation amount of HINT is also less than that of TextGCN. The performance of our model is not only better than those of other models, but the numbers of parameters and calculations are also very small, which further proves that our model is simple and effective. 

For text classification or other NLP tasks, increasing the complexity of the neural network model and the number of network parameters can often achieve better performance. Therefore, the problem is that the development of NLP tasks highly depends on computing power. Our model achieves the improvement of model performance while reducing the complexity of the model. HINT makes the extraction of features not completely dependent on the deep learning network and greatly reduces the requirements for the computing power of the neural network.

\section{Conclusion}
In this paper, we proposed a novel method to address the limitation of previous works in text hierarchical information utilization. We build a dependency parsing graph for each text and construct a coding tree for each graph by structural entropy minimization algorithm. Our model uses the hierarchy of the coding tree to learn the representation of each text. Experimental results demonstrate the ability of HINT to decode hierarchical information in text and show the positive effect of hierarchical information on text classification tasks. Our model achieves state-of-the-art performance with a simple structure and few parameters.

\section*{Acknowledgements}
This research was supported by NSFC (Grant No. 61932002).

\bibliography{reference}
\bibliographystyle{acl_natbib}

\end{document}